\documentclass[runningheads]{llncs}
\usepackage{amsmath}
\usepackage{amssymb}
\usepackage{epsfig}
\usepackage{graphicx}
\usepackage{pifont}
\usepackage{textcomp}
\usepackage{times}
\usepackage{mathtools}
\usepackage{mathrsfs}
\usepackage{subcaption}
\usepackage{multirow}
\usepackage{listings}
\usepackage[normalem]{ulem}
\usepackage[svgnames]{xcolor}
\usepackage{xspace}
\newcommand\norm[1]{\left\lVert#1\right\rVert}

\newcommand\YAMLcolonstyle{\color{red}\mdseries}
\newcommand\YAMLkeystyle{\color{black}\bfseries}
\newcommand\YAMLvaluestyle{\color{blue}\mdseries}
\lstdefinelanguage{yaml}{
    keywords={true,false,null,y,n},
    keywordstyle=\color{darkgray}\bfseries,
    basicstyle=\YAMLkeystyle,
    sensitive=false,
    comment=[l]{\#},
    morecomment=[s]{/*}{*/},
    commentstyle=\color{blue}\ttfamily,
    stringstyle=\YAMLvaluestyle,
    moredelim=[l][\color{orange}]{\&},
    moredelim=[l][\color{blue}\mdseries]{*},
    moredelim=**[il][\YAMLcolonstyle{:}\YAMLvaluestyle]{:},
    morestring=[b]',
    morestring=[b]"
}

\newcommand{\FRAMEWORK}{{\bf torchdistill}\xspace}

\begin{document}
%
\title{\textsf{torchdistill}: A Modular, Configuration-Driven Framework for Knowledge Distillation\thanks{Accepted to the 3rd Workshop on Reproducible Research in Pattern Recognition at ICPR 2020.}}
%
%
\author{Yoshitomo Matsubara\inst{1}\orcidID{0000-0002-5620-0760}}
\authorrunning{Y. Matsubara}
%
\institute{
University of California, Irvine, CA 92697, USA
\email{yoshitom@uci.edu}
}
\maketitle              
\begin{abstract}
While knowledge distillation (transfer) has been attracting attentions from the research community, the recent development in the fields has heightened the need for reproducible studies and highly generalized frameworks to lower barriers to such high-quality, reproducible deep learning research.
Several researchers voluntarily published frameworks used in their knowledge distillation studies to help other interested researchers reproduce their original work.
Such frameworks, however, are usually neither well generalized nor maintained, thus researchers are still required to write a lot of code to refactor/build on the frameworks for introducing new methods, models, datasets and designing experiments.
In this paper, we present our developed open-source framework built on PyTorch and dedicated for knowledge distillation studies.
The framework is designed to enable users to design experiments by declarative PyYAML configuration files, and helps researchers complete the recently proposed ML Code Completeness Checklist.
Using the developed framework, we demonstrate its various efficient training strategies, and implement a variety of knowledge distillation methods.
We also reproduce some of their original experimental results on the ImageNet and COCO datasets presented at major machine learning conferences such as ICLR, NeurIPS, CVPR and ECCV, including recent state-of-the-art methods.
All the source code, configurations, log files and trained model weights are publicly available at \url{https://github.com/yoshitomo-matsubara/torchdistill}.

\keywords{Knowledge distillation \and Open source framework \and Reproducibility.}
\end{abstract}
\section{Introduction}
\label{sec:intro}

Deep learning methods have been achieving state-of-the-art performances, contributing to the rapid development of applications for a variety of tasks such as image classification~\cite{he2016deep,mahajan2018exploring,tan2019efficientnet,touvron2019fixing} and object detection~\cite{ren2015faster,he2017mask,carion2020end}. 
One of the critical problems with such state-of-the-art models is their complexity, thus the complex models are difficult to be deployed for real-world applications.
In general, there is a trade-off between model complexity and inference performance (\emph{e.g.}, measured as accuracy), and there are three different types of method to make models deployable: 1) designing lightweight models, 2) model compression/pruning, and 3) knowledge distillation.
Lightweight models such as MobileNet~\cite{sandler2018mobilenetv2,howard2019searching}, MnasNet~\cite{tan2019mnasnet} and YOLO series~\cite{redmon2017yolo9000,redmon2018yolov3} often sacrifice inference performance to reduce inference time, compared to complex models \emph{e.g.}, ResNet~\cite{he2016deep} and Mask R-CNN~\cite{he2017mask}.
Model compression and pruning~\cite{han2016deep,li2016pruning} techniques reduce model size by quantizing parameters and pruning redundant neurons, and such methods are covered by Distiller~\cite{zmora2019neural}, an open-source library for model compression.

In this paper, our focus is on the last category, knowledge distillation, that trains a simpler (student) model to mimic the behavior of a powerful (teacher) model.
Knowledge distillation~\cite{hinton14distilling} stems from the study by Bucilu\v{a}~\emph{et al.}~\cite{bucilua2006model}, that presents a method to compress large, complex ensembles into smaller models with small loss in inference performance.
Interestingly, Ba and Caruana~\cite{ba2014deep} report that student models trained to mimic the behavior of the teacher models (\emph{soft-label}) significantly outperform those trained on the original (\emph{hard-label}) dataset.
Following these studies, knowledge distillation and transfer have been attracting attention from the research communities such as computer vision~\cite{romero2015fitnets} and natural language processing~\cite{sanh2019distilbert}.

\begin{table}[t]
    \caption{Knowledge distillation frameworks. \FRAMEWORK supports modules in PyTorch and torchvision such as loss, datasets and models. ImageNet: ILSVRC 2012~\cite{russakovsky2015imagenet}, YT Faces: YouTube Faces DB~\cite{wolf2011face}, MIT Scenes: Indoor Scenes dataset~\cite{quattoni2009recognizing}, CUB-2011: Caltech-UCSD Birds-200-2011~\cite{wah2011caltech}, Cars: Cars dataset~\cite{krause20133d}, SOP: Stanford Online Products~\cite{oh2016deep}. P: Pretrained models, M: Module abstraction, D: Distributed training.}
    \vspace{-0.5em}
    \begin{center}
    \bgroup
    \setlength{\tabcolsep}{0.5em}
    \def\arraystretch{1.1}
        \begin{tabular}{|c|c|c|c|c|c|}
            \hline
            \textbf{Framework} & \textbf{Supported datasets} & \textbf{Models} & \textbf{P} & \textbf{M} & \textbf{D}\\ \hline \hline
            Zagoruyko \& Komodakis~\cite{zagoruyko2017paying} & CIFAR-10, ImageNet & Hard-coded & \checkmark & & \\
            Passalis \& Tefas~\cite{passalis2018learning} & CIFAR-10, YT Faces & Hard-coded & & & \\
            Heo~\emph{et al.}~\cite{heo2019knowledge} & CIFAR-10, MIT scenes & Hard-coded & \checkmark & & \\
            Park~\emph{et al.}~\cite{park2019relational} & Cars, CUB-2011, SOP & Hard-coded & & & \\
            Tian~\emph{et al.}~\cite{tian2020contrastive} & CIFAR-100 & Hard-coded & \checkmark & & \\
            Yuan~\emph{et al.}~\cite{yuan2020revisiting} & CIFAR-10, -100, Tiny ImageNet & Hard-coded & \checkmark & & \\
            Xu~\emph{et al.}~\cite{xu2020knowledge} & CIFAR-100 & Hard-coded & \checkmark & & \\
            \FRAMEWORK & torchvision* & torchvision* & \checkmark & \checkmark & \checkmark \\
            \hline
        \end{tabular}
    \egroup
    \end{center}
    \vspace{-0.5em}
    \small * \FRAMEWORK supports those implemented with PyTorch. In this paper, our focus is on torchvision. 
    \label{table:related_work}
\end{table}

As summarized in Table~\ref{table:related_work}, some researchers voluntarily publish their knowledge distillation frameworks \emph{e.g.},~\cite{zagoruyko2017paying,passalis2018learning,heo2019knowledge,park2019relational,tian2020contrastive,xu2020knowledge} to help other researchers reproduce their original studies.
However, such frameworks are usually not either well generalized or maintained to be built on.
Besides, Distiller~\cite{zmora2019neural} supports only one method for knowledge distillation, and Catalyst~\cite{catalyst} is a framework built on PyTorch with a focus on reproducibility of deep learning research.
To support various deep learning methods, these frameworks are well generalized, yet require users to \emph{hardcode} (reimplement) critical modules such as models and datasets, even if the implementations are publicly available in popular libraries, to design complex knowledge distillation experiments.
As pointed out by Gardner~\emph{et al.}~\cite{gardner2018allennlp}, reference methods and models are often re-implemented from scratch, and this makes it difficult to reproduce the reported results.
For further advancing the deep learning research, a new generalized framework is therefore needed, and the framework should be able to allow researchers to easily try different modules (\emph{e.g.}, models, datasets, loss configurations), implement various approaches, and take care of reproducibility of their work. 

The concept of our framework, \FRAMEWORK,\footnote{\label{fn:our_repo}\url{https://github.com/yoshitomo-matsubara/torchdistill}} is highly inspired by AllenNLP~\cite{gardner2018allennlp}, a platform built on PyTorch~\cite{paszke2019pytorch} for research on deep learning methods in natural language processing.
Similar to AllenNLP, \FRAMEWORK supports the following features:
\begin{itemize}
    \setlength\itemsep{0em}
    \item module abstractions that enable researchers to write higher-level code for experiments \emph{e.g.}, model, dataset, optimizer and loss;
    \item declarative PyYAML configuration files, which can be seen as high-level summaries of experiments (training and evaluation), enable to use anchors and aliases in the file to refer to the same object (\emph{e.g.}, file paths) and simplify themselves, and make it easy to change the abstracted components and hyper-parameters; and
    \item generalized reference code and configurations to apply knowledge distillation methods to PyTorch and torchvision models pretrained on well-known complex benchmark datasets: ImageNet (ILSVRC 2012)~\cite{russakovsky2015imagenet} and COCO 2017~\cite{lin2014microsoft}.
\end{itemize}

Furthermore, \FRAMEWORK supports 1) seamless multi-stage training, 2) caching teacher's outputs, and 3) redesigning (pruning) teacher and student models without hard-coding  (reimplementation).
To the best of our knowledge, this is the first, highly generalized open-source framework that can support a variety of knowledge distillation methods, and lower barriers to high-quality, reproducible deep learning research~\cite{gundersen2018state}.
Researchers can explore methods and shape new approaches, building on this generalized framework that makes it easy not only to customize existing methods and models, but also introduce completely new ones.
Using some of our reimplemented methods, we also reproduce the experimental results on ILSVRC 2012 and COCO 2017 datasets reported in the original studies.

\section{Framework Design}
\label{sec:design}
Our developed framework, \FRAMEWORK, is an open source framework dedicated for knowledge distillation studies, built on PyTorch~\cite{paszke2019pytorch}.
For vision tasks such as image classification and object detection, the framework is designed to support torchvision, that offers a lot of options for datasets, model architectures and common image transformations.
The collection of supported reference models and datasets in our framework are dependent on the version of user's installed torchvision.
For instance, when users find new models in the latest torchvision, they can shortly try the models simply by updating the torchvision and configuration files for their experiments with our framework.

\subsection{Module Abstractions}
\label{subsec:abstraction}

An objective of module abstractions in our framework is to enable researchers to experiment with various modules by simply changing a PyYAML configuration file described in Section~\ref{subsec:configs}.
We focus abstraction on critical modules to experiment, specifically model architectures, datasets, transforms, and losses to be minimized during training.
These modules are often hard-coded (See Appendix~\ref{appendix:hardcoded}) in authors' published frameworks~\cite{zagoruyko2017paying,passalis2018learning,heo2019knowledge,park2019relational,tian2020contrastive,xu2020knowledge}, and many of the hyperparameters are hard-coded as well.

\paragraph{Model architectures:}
torchvision offers various model families for vision tasks from AlexNet~\cite{krizhevsky2012imagenet} to R-CNNs~\cite{ren2015faster,he2017mask}, and many of them are pretrained on large benchmark datasets.
Specifically, the latest release (v0.8.2) provides about 30 image classification models pretrained on ImageNet (ILSVRC 2012)~\cite{russakovsky2015imagenet} and 4 object detection models pretrained on COCO 2017~\cite{lin2014microsoft}.
As our framework supports torchvision for vision tasks, researchers can use such pretrained models as teacher and/or baseline models (\emph{e.g.}, student trained without teacher).
In addition to the pretrained models available in torchvision, they can use their own pretrained model weights and any model architectures implemented with PyTorch.
Moreover, \FRAMEWORK supports PyTorch Hub\footnote{\url{https://pytorch.org/hub/}} and enable users to import modules via the hub by specifying repository names in a PyYAML configuration file.

\paragraph{Datasets:}
As described above, torchvision also supports a variety of datasets, and previous studies~\cite{romero2015fitnets,yim2017gift,zagoruyko2017paying,passalis2018learning,kim2018paraphrasing,heo2019knowledge,park2019relational,wang2019distilling,ahn2019variational,matsubara2019distilled,peng2019correlation,tung2019similarity,tian2020contrastive} use many of them to validate proposed distillation techniques such as ImageNet~\cite{russakovsky2015imagenet}, COCO~\cite{lin2014microsoft}, CIFAR-10 and -100~\cite{krizhevsky2009learning}, and Caltech101~\cite{fei2006one}.
Similar to model architectures, \FRAMEWORK supports such datasets and can collaborate with any datasets implemented with PyTorch.

\paragraph{Transforms:}
In vision tasks, there are de facto standard image transform techniques.
Taking image classification on the ImageNet dataset as an example, a standard transform pipeline for training with torchvision\footnote{\label{fn:torchvision_train}\url{https://github.com/pytorch/vision/blob/master/references/classification/train.py}} consists of 1) making a crop of random size of the original size and  with a random aspect ratio of the original aspect ratio, 2) horizontal reflection with 50\% chance for data augmentation to reduce a risk of overfitting~\cite{krizhevsky2012imagenet}, 3) PIL-to-Tensor conversion, and 4) channel-wise normalization using (0.485, 0.456, 0.406) and (0.229, 0.224, 0.225) as means and standard deviations, respectively.
In \FRAMEWORK, users can define their own transform pipeline in a configuration file.

\paragraph{Losses:}
In distillation process, student models are trained using outputs from teacher models, and the research community has been proposing a lot of unique losses with/without task-specific losses such as cross entropy loss for classification tasks.
PyTorch~\cite{paszke2019pytorch} supports various loss classes/functions, and simple distillation losses can be defined in a configuration file by combining such supported losses using \FRAMEWORK's customizable loss module (See Section~\ref{subsec:custom_loss_module}).

\subsection{Registry}
\label{subsec:registry}

The registry is an important component in \FRAMEWORK as abstracted modules are instantiated by mapping strings in the configuration file to the objects in code.
Furthermore, it would make it easy for users to collaborate their implemented modules/functions with this framework.
Similar to AllenNLP~\cite{gardner2018allennlp} and Catalyst~\cite{catalyst}, this can be done even outside the framework by using a Python decorator.
The following example shows that a new model class, \emph{MyModel}, is added to the framework by simply using \emph{@register\_model} (defined in the framework), and the new class can be instantiated by defining ``MyModel'' with required parameters at designated places in a configuration file.

\begin{lstlisting}[language=Python, backgroundcolor=\color{MintCream}, columns=fullflexible]
@register_model
class MyModel(nn.Module):
    def __init__(self, *args, **kwargs):
        super().__init__()
        self.conv1 = nn.Conv2d(**kwargs['conv1_kwargs'])
        ...
\end{lstlisting}

\subsection{Configurations}
\label{subsec:configs}
An experiment can be defined by a PyYAML configuration file (See Appendix~\ref{appendix:example_config}), that allows users to tune hyperparameters, and change methods/models without hard-coding.
With PyYAML's features, configuration files allow users to leverage anchors and aliases, and these features would be helpful to simplify the configurations in cases that users would like to reuse parameters defined in the configuration file such as root directory path for datasets, parameters and model names as part of checkpoint file paths for better data management.
In a configuration file, there are three main components to be defined: datasets, teacher and student models, and training.
Each of the key components is defined by using abstracted and registered modules described in Sections~\ref{subsec:abstraction} and \ref{subsec:registry}.
A configuration file gives users a summary of the experiment, and shows all the parameters to reproduce the experimental results except implicit factors such as hardware specifications used for the experiment.

The following example illustrates how to define a global teacher model declared in a PyYAML configuration file.
As described in the previous sections, various types of modules are abstracted in our framework, and such modules (classes and functions) in user's installed torchvision are registered.
In this example, 'resnet34' function\footnote{\url{https://pytorch.org/docs/stable/torchvision/models.html\#torchvision.models.resnet34}} is used to instantiate an object of type \emph{ResNet} by using a dictionary of keyword arguments (\emph{**params}). \emph{i.e.} \emph{num\_classess} = 1000 and \emph{pretrained} = True are given as arguments of 'resnet34' function.
For image classification models implemented in torchvision or those users add to the registry in our framework, users can easily try different models by changing 'resnet34' \emph{e.g.}, 'densenet201'~\cite{huang2017densely}, 'mnasnet1\_0'~\cite{tan2019mnasnet}.
Besides that, \emph{ckpt} indicates the file path of checkpoint, that is './resnet34.pt' in the example defined by leveraging some of YAML features: anchors (\&) and aliases (*).
For teacher model, the checkpoint will be used to initialize the model with user's own model weights if the checkpoint file exists.
Otherwise, 'resnet34' in this example will be initialized with torchvision's pretrained weights for ILSVRC 2012.

\begin{lstlisting}[language=yaml, backgroundcolor=\color{MintCream}]
teacher_model:
    name: &teacher 'resnet34'
    params:
        num_classes: 1000
        pretrained: True
    ckpt: !join ['./', *teacher, '.pt']
\end{lstlisting}

Furthermore, \FRAMEWORK offers an option to generate log files that monitor the experiments.
For instance, a log file presents what parameters were used, when executed, the trends of training behavior (\emph{e.g.}, training loss, learning rate and validation accuracy) at a frequency set in the configuration file, and evaluation results.

These configuration and log files\footnote{\label{fn:configs}Available at \url{https://github.com/yoshitomo-matsubara/torchdistill/tree/master/configs/}.} will also help the researchers complete \emph{ML Code Completeness Checklist},\footnote{\label{fn:ml_code_checklist}\url{https://github.com/paperswithcode/releasing-research-code}} that was recently proposed to facilitate reproducibility in the research community as part of the official code submission process at major machine learning conferences \emph{e.g.}, NeurIPS, ICML and CVPR.

\subsection{Dataset Wrappers}
\label{subsec:dataset_wrappers}

To support a wide variety of knowledge distillation methods, dataset is an important module to be generalized.
Usually, the dataset module in PyTorch and torchvision returns a pair of input batch (\emph{e.g.}, collated image tensors) and targets (ground-truth) at each iteration, but some of the existing knowledge distillation approaches require additional information for the batch.
For instance, contrastive representation distillation (CRD)~\cite{tian2020contrastive} requires an efficient strategy to retrieve a large number of negative samples in the training session, that requires the dataset module to return an additional object (\emph{e.g.}, negative sample indices).
To support such extensions, we design dataset wrappers to return input batch, targets, and a supplementary dictionary, that can be empty when not used.
For the above case, the additional object can be stored in the supplementary dictionary, and used when computing the contrastive loss.
This design also enables us to support caching teacher model's outputs against data indices in the original dataset so that teacher's inference can be skipped by caching (serializing) outputs of the teacher model given a data index at the first epoch, and reading and collating the cached outputs given batch of data indices at the following epochs.

\begin{figure}[t]
    \centering
    \begin{subfigure}[b]{0.49\linewidth}
        \centering
        \includegraphics[width=1.0\linewidth]{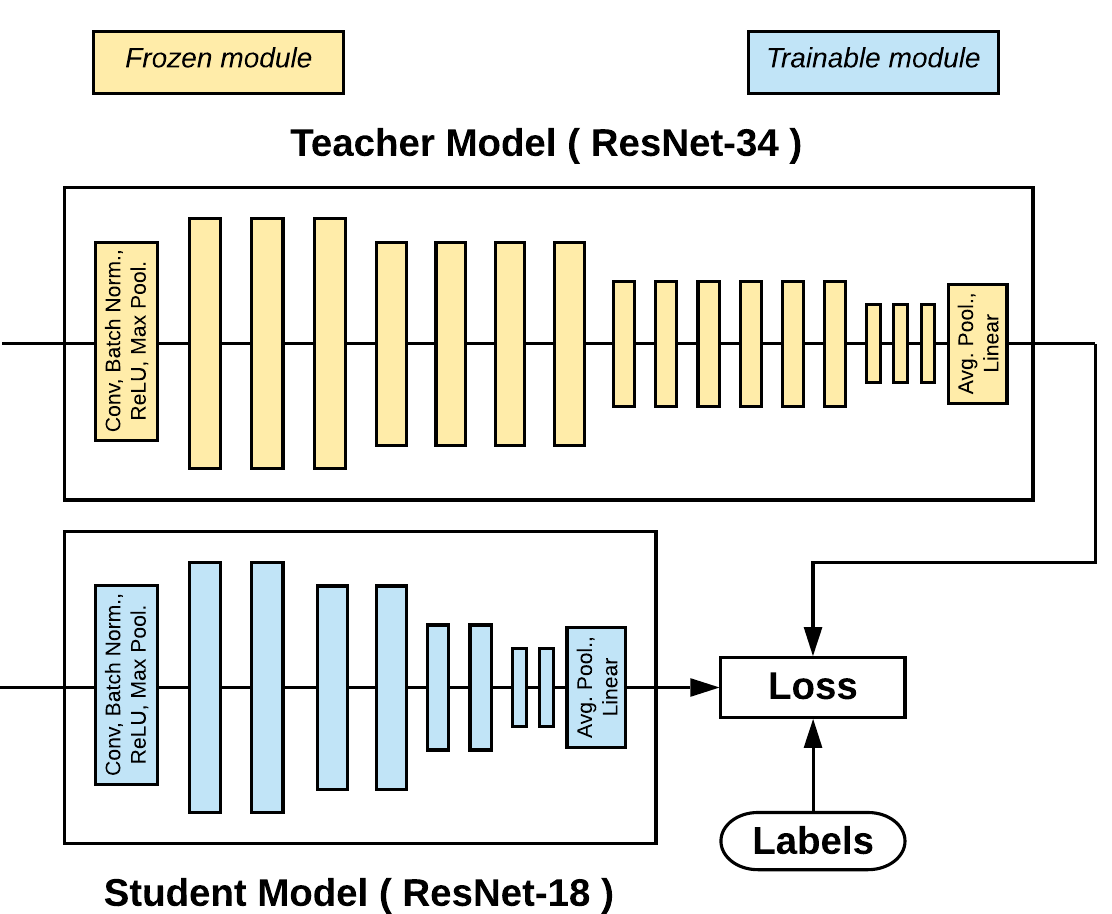}
        \caption{Knowledge distillation~\cite{hinton14distilling} using ResNet-34 and ResNet-18 as teacher and student models, respectively.\vspace{1em}}
        \label{fig:kd}
    \end{subfigure}
    \begin{subfigure}[b]{0.49\linewidth}
        \centering 
        \includegraphics[width=1.0\linewidth]{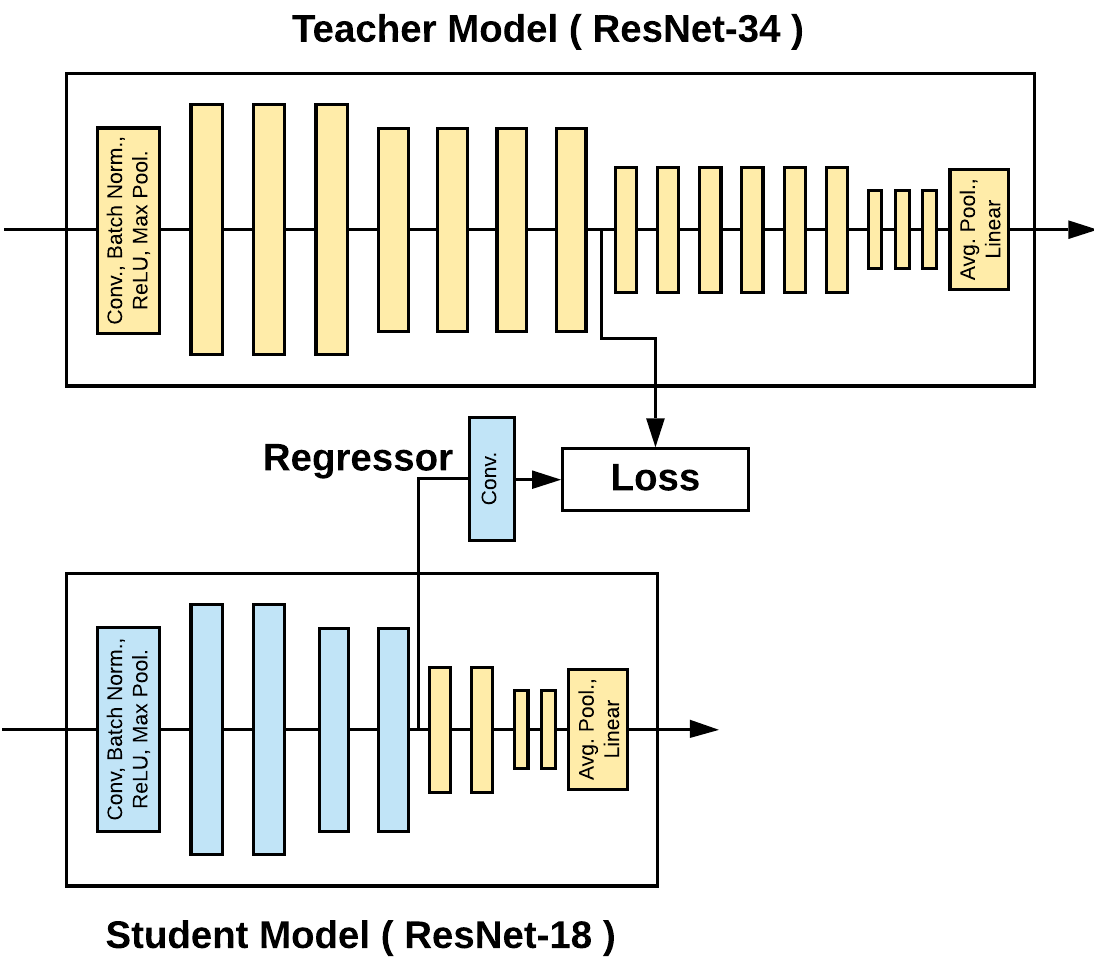}
        \caption{Hint-training with an auxiliary module (convolutional regressor) as stage 1 of FitNet method~\cite{romero2015fitnets}. Its stage 2 is knowledge distillation as illustrated in Figure~\ref{fig:kd}.}
    \end{subfigure}
    \vspace{-1em}
    \caption{Knowledge distillation and FitNet methods. Yellow and blue modules indicate that their parameters are frozen and trainable, respectively.}
    \label{fig:hint_training}
\end{figure}

To demonstrate that caching improves training efficiency, we perform an experiment with knowledge distillation~\cite{hinton14distilling} illustrated in Fig.~\ref{fig:kd} that caches outputs of the teacher model at the first epoch for training ResNet-18 (student) on ILSVRC 2012 dataset, and skips the teacher model's inference by loading and feeding the outputs cached on disk to the loss module.
Table~\ref{table:cacheable_dataset} suggests that spending an extra one-minute at the 1st epoch to serialize teacher's outputs, the caching strategy makes the following training process (\emph{i.e.} from the 2nd epoch) approximately 1.23 -- 2.11 times faster at epoch-level when using 3 NVIDIA GeForce RTX 2080 Ti`s with batch size of 256.
Also, this improvement becomes more significant when using a larger teacher model such as ResNet-152 (approximately 2.11 times faster than training without cache).
The ILSVRC 2012 training dataset consists of approximately 1.3 million images, and the cached files consumes only 10GB whereas the original training dataset uses about 140GB.
Note that caching may not improve the training efficiency if teacher's outputs to be cached are much larger \emph{e.g.}, hint-based training~\cite{romero2015fitnets} requires intermediate outputs from teacher and student models.
Also, this mode should be turned off when applying data augmentation strategies.

\begin{table}[t]
    \caption{Epoch-level training speed improvement by \underline{\textbf{caching teacher's outputs}} at the 1st epoch, using ResNet-18 as student model for knowledge distillation~\cite{hinton14distilling}.}
    \begin{center}
    \bgroup
    \setlength{\tabcolsep}{0.25em}
    \def\arraystretch{1.1}
    \small
        \begin{tabular}{|c|r|r|r|r|}
            \hline
            \textbf{Teacher} & \textbf{ResNet-34} & \textbf{ResNet-50} & \textbf{ResNet-101} & \textbf{ResNet-152} \\ \hline
            No cache & 801 sec & 1,030 sec & 1,348 sec & 1,944 sec \\
            Cache (1st) & 859 sec & 1,079 sec & 1,402 sec & 1,966 sec \\
            Cache (2nd) & {\bf 651 sec} & {\bf 649 sec} & {\bf 656 sec} & {\bf 917 sec} \\
            \hline
        \end{tabular}
    \egroup
    \end{center}
    \label{table:cacheable_dataset}
\end{table}

\subsection{Teacher and Student Models}
\label{subsec:teacher_student}
\begin{figure}[t]
    \centering
    \begin{subfigure}[b]{0.8\linewidth}
        \centering
        \includegraphics[width=1.0\linewidth]{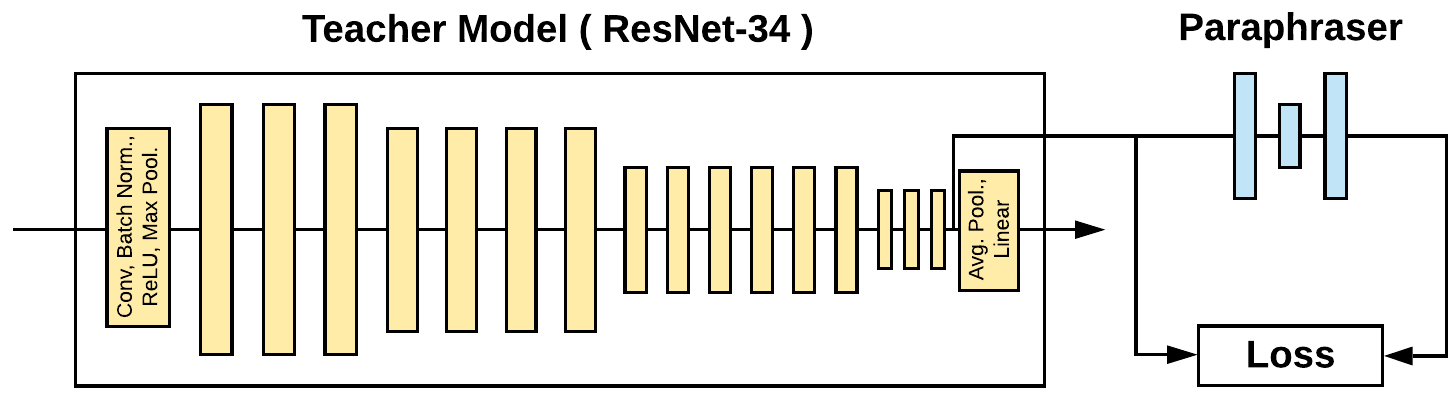}
        \caption{\underline{1st stage}: training \emph{paraphraser} for teacher model.}
        \label{fig:ft_1st}
    \end{subfigure}
    \\ \vspace{1em}
    \begin{subfigure}[b]{0.8\linewidth}
        \centering 
        \includegraphics[width=1.0\linewidth]{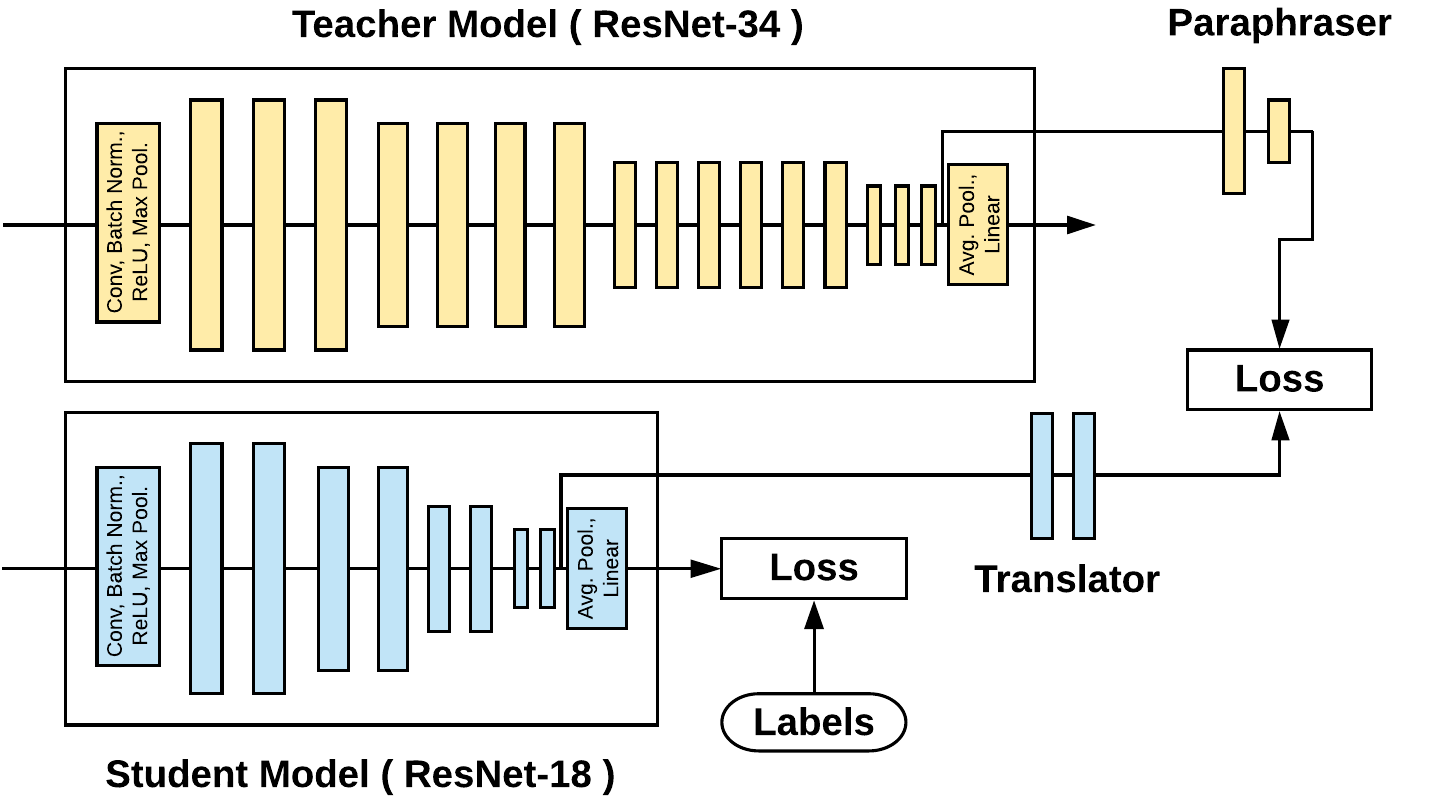}
        \caption{\underline{2nd stage}: training student model and \emph{translator}, using labels and outputs of \emph{paraphraser}'s middle layer.}
        \label{fig:ft_2nd}
    \end{subfigure}
    \vspace{-1em}
    \caption{Factor transfer with two auxiliary modules.} 
    \label{fig:ft}
\end{figure}

Teacher-Student pairs are keys in knowledge distillation experiments, and recently proposed approaches~\cite{romero2015fitnets,yim2017gift,zagoruyko2017paying,heo2019knowledge,ahn2019variational,peng2019correlation,tian2020contrastive,xu2020knowledge,zhang2020prime} introduce auxiliary modules, which are used only in training session.
Such auxiliary modules use tensors from intermediate layers in models, and introducing the modules to the models often results in branching their feedforward path as shown in Figs.~\ref{fig:hint_training} and \ref{fig:ft}.
This paradigm, however, is also one of the backgrounds that researchers decide to hard-code the models (\emph{e.g.}, modify the original implementations of models in torchvision every time they change the placement of auxiliary modules for preliminary experiments) to introduce such auxiliary modules used for their proposed methods, and make it difficult for other researchers to build on the published frameworks~\cite{zagoruyko2017paying,passalis2018learning,heo2019knowledge,park2019relational,tian2020contrastive,xu2020knowledge}.

Taking an advantage of forward hook paradigm in PyTorch~\cite{paszke2019pytorch}, \FRAMEWORK supports introducing such auxiliary modules without altering the original implementations of the models.
Specifically, users can register the framework's provided \emph{forward hooks} to specific modules to store its input and/or output in a I/O dictionary by specifying the module paths (\emph{e.g.}, ``conv1'' for a MyModel object in Section~\ref{subsec:registry}) in the configuration files.
The I/O dictionaries for teacher and student models will be fed to a generalized, customizable loss module described in Section~\ref{subsec:custom_loss_module}.

For methods that not only require to extract the intermediate outputs (See Fig.~\ref{fig:hint_training}) but also feed the extracted outputs to trainable auxiliary modules in different branches to be processed (See Fig.~\ref{fig:ft_2nd}), we define a \emph{special} module in the framework, that is designed to have a \emph{post-forward} function.
In Fig.~\ref{fig:hint_training}, for instance, the framework first executes ResNet-18 and extracts intermediate output by a registered forward hook, and then the extracted output stored in the student's I/O dictionary will be fed to the regressor as part of the \emph{post-forward} process.
The concept of the special module gives users more flexibility in designing training methods while leaving the original implementations of models (ResNet-34 and ResNet-18 in Fig.~\ref{fig:ft}) unaltered.

\subsection{Customizable Loss Module}
\label{subsec:custom_loss_module}
Leveraging the I/O dictionaries that contain input/output of specific modules with registered forward hooks, \FRAMEWORK provides a generalized customizable loss module that allows users to easily combine different loss modules with balancing factors by configuration files such as those in Fig.~\ref{fig:ft_2nd}.
Given a pair of input $x$ and ground-truth $y$, the I/O dictionaries consist of a set of keys $J$ and the values $z_{j}^{\textrm S}$ and $z_{j}^{\textrm T}$ ($j \in J$) extracted from student and teacher models respectively.
Using the I/O dictionaries and the ground-truth, the generalized loss is defined as

\begin{equation}
    \mathcal{L} = \sum_{j \in J} \lambda_{j} \cdot \mathcal{L}_{j}(z_{j}^{\textrm S}, z_{j}^{\textrm T}, y),
    \label{eq:custom_loss}
\end{equation}

\noindent where $\lambda_{j}$ is a balancing weight (hyperparameter) for $\mathcal{L}_{j}$, which is either a loss module implemented in PyTorch~\cite{paszke2019pytorch} or user's defined loss module in registry.

For instance, the loss function to train student model on ILSVRC 2015 dataset~\cite{russakovsky2015imagenet} at the 2nd stage of factor transfer (Fig.~\ref{fig:ft_2nd}) can be defined as:

\begin{align}
    \mathcal{L} = \lambda_{\textrm{cls}} &\cdot \mathcal{L}_{\textrm{cls}}(z_{\textrm{cls}}^{\textrm S}, z_{\textrm{cls}}^{\textrm T}, y) + \lambda_{\textrm{FT}} \cdot \mathcal{L}_{\textrm{FT}}(z_{\textrm{FT}}^{\textrm S}, z_{\textrm{FT}}^{\textrm T}, y)\\
    \nonumber
    &\mathcal{L}_{\textrm{cls}}(z_{\textrm{cls}}^{\textrm S}, z_{\textrm{cls}}^{\textrm T}, y) = \textrm{CrossEntropyLoss}(z_{\textrm{cls}}^{\textrm S}, y) \\ \nonumber
    &\mathcal{L}_{\textrm{FT}}(z_{\textrm{FT}}^{\textrm S}, z_{\textrm{FT}}^{\textrm T}, y) = \norm{\frac{z_{\textrm{FT}}^{\textrm S}}{\norm{z_{\textrm{FT}}^{\textrm S}}_{2}} - \frac{z_{\textrm{FT}}^{\textrm T}}{\norm{z_{\textrm{FT}}^{\textrm T}}_{2}}}_{p},
\end{align}
\noindent where $\lambda_{\textrm{cls}} = 1$, $\lambda_{\textrm{FT}} = 1,000$ and $p$ = 1, following~\cite{kim2018paraphrasing}.

\subsection{Stage-wise Training Configuration}
In the previous sections, we describe the main features of \FRAMEWORK, and what modules are configurable in the framework.
We emphasize that all the training configurations described above can be defined stage-wisely.

\paragraph{Seamless multi-stage training configurations:}
Specifically, the framework is designed to enable users to configure critical components such as 1) number of epochs, 2) training and validation datasets, 3) teacher and student models, 4) modules (layers) to be trained/frozen, 5) optimizer, 6) learning rate scheduler, 7) loss module.
These components can be re-defined at each of training stages, otherwise the framework reuses those from the previous stage.
Notice that these training configurations can be declared in a configuration file, and this design enables to support not only two-stage training strategies~\cite{romero2015fitnets,yim2017gift,kim2018paraphrasing,heo2019knowledge}, but also more complicated distillation methods such as teacher assistant knowledge distillation (TAKD)~\cite{mirzadeh2019improved}, that trains TAs to fill the gap between student and teacher models.
Transfer learning also can be supported by changing models and datasets from stage to stage, and users would execute code with a configuration file only once.
Therefore, they will not need to execute code multiple times to perform multi-stage training, including transfer learning.

\begin{figure}[t]
    \centering
    \includegraphics[width=0.55\linewidth]{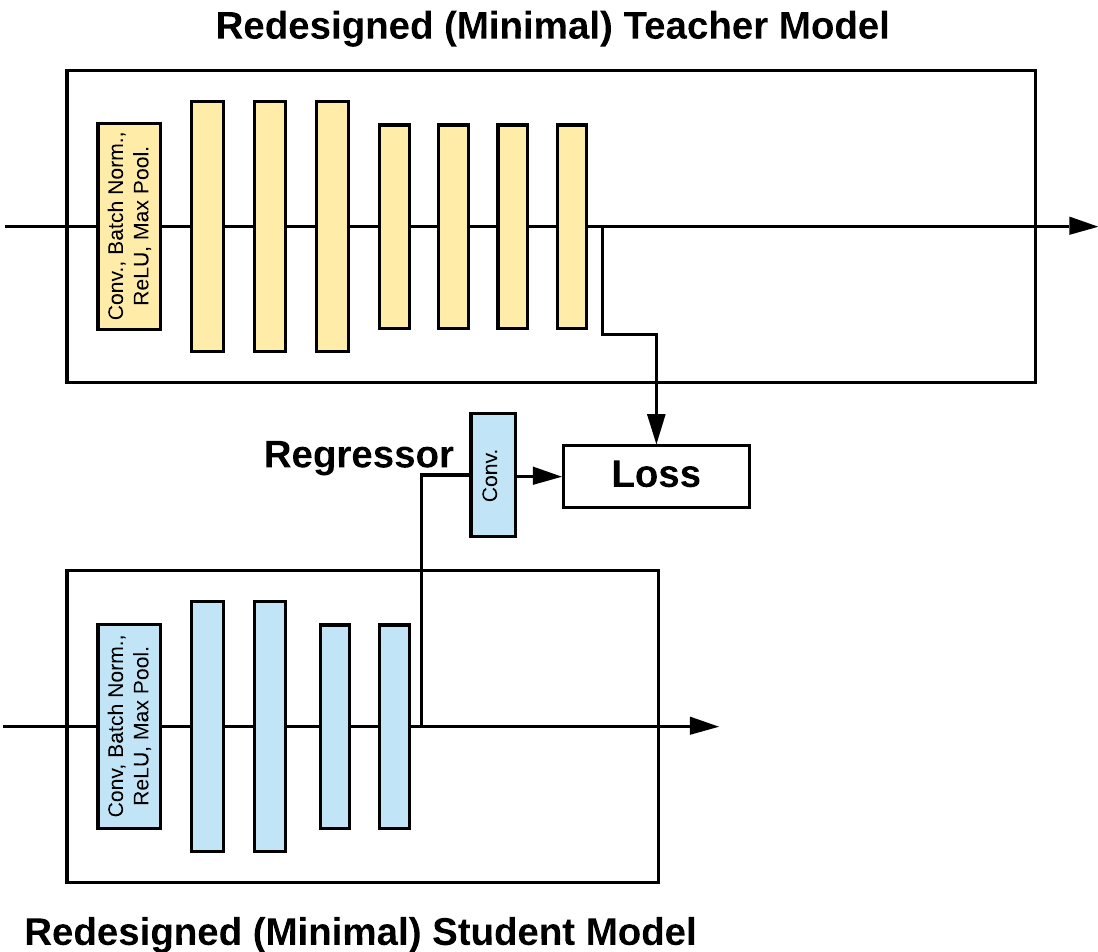}
    \caption{Hint-training with teacher and student models pruned simply by specifying required modules in a configuration file for further efficient training, compared to a naive configuration in Fig.~\ref{fig:hint_training}.}
    \label{fig:min_hint_training}
\end{figure}

\paragraph{Redesigning models for efficient training:}
Furthermore, our framework gives users an option to redesign teacher and student models at each stage by specifying the required modules in a configuration file.
Specifically, users are allowed to rebuild models by reusing modules in the models optionally with auxiliary modules.
Figure~\ref{fig:hint_training} shows an example that modules after the 8th and the 5th blocks of the teacher and student models respectively can be pruned as the outputs of the modules are not used in the hint-training (1st stage), thus not required to be executed.
In this specific case, the redesigned student model will consist of the trainable (blue) modules and a regressor (auxiliary module) as illustrated in Fig.~\ref{fig:min_hint_training}, and the teacher and student architectures at the 2nd stage will be reverted to the original ones (Fig.~\ref{fig:kd}) with parameters learnt at the 1st stage.
Also, the redesigned teacher/student model can be an empty module to save execution time.
In Fig.~\ref{fig:ft_1st}, for instance, there is no need to feed input batch to the student model (thus, can be empty) as at the 1st stage of factor transfer, only the teacher model is executed to train the paraphraser.

As introduced in Section~\ref{subsec:dataset_wrappers}, when the teacher's outputs are cacheable (\emph{e.g.}, in terms of available disk space), teacher's inference can be skipped by loading the cache files produced at previous epoch.
Redesigning models help users shorten training sessions even when teacher's outputs are not cacheable.
Note that student model's outputs, however, cannot be cached as the model's parameters are updated every iteration.
Table~\ref{table:model_redesign} suggests that redesigning models using only modules to be executed for training would be an effective approach to saving training time, and this improvement would be more critical for training models on large datasets and/or with a lot of epochs.
We emphasize that users can redesign (minimize) the models by specifying the required modules in a configuration file rather than hardcode (reimplement) the pruned models.

\begin{table}[t]
    \caption{Epoch-level training speed improvement by \textbf{redesigning teacher and student (ResNet-18) models} with required modules only for hint-training shown in Figure~\ref{fig:min_hint_training}.}
    \begin{center}
    \bgroup
    \setlength{\tabcolsep}{0.3em}
    \def\arraystretch{1.1}
    \small
        \begin{tabular}{|c|r|r|r|r|}
            \hline
            \textbf{Teacher} & \textbf{ResNet-34} & \textbf{ResNet-50} & \textbf{ResNet-101} & \textbf{ResNet-152} \\ \hline
            Original & 934 sec & 1,175 sec & 1,468 sec & 1,779 sec \\
            Minimal & {\bf 786 sec} & {\bf 929 sec} & {\bf 936 sec} & {\bf 1,022 sec} \\
            \hline
        \end{tabular}
    \egroup
    \end{center}
    \label{table:model_redesign}
\end{table}

\section{Reference Methods}
\label{sec:ref_methods}
Here, we describe the reimplementations of knowledge distillation methods and experiments to reproduce the reported results on ImageNet and COCO datasets.

\begin{table}[t]
    \caption{Reference knowledge distillation methods implemented in \FRAMEWORK.}
    \begin{center}
    \bgroup
    \setlength{\tabcolsep}{0.3em}
    \def\arraystretch{1.1}
    \small
        \begin{tabular}{|c|c|c|c|c|c|}
            \hline
            \multirow{2}{*}{\bf Methods} & \bf Multi-stage & \multicolumn{3}{c|}{\bf Required additional modules} \\ \cline{3-5}
             & \bf training & Auxiliary & Special & Custom dataset\\ \hline
            KD~\cite{hinton14distilling} & & & & \\
            FitNet~\cite{romero2015fitnets} & \checkmark & \checkmark & & \\
            FSP~\cite{yim2017gift} & \checkmark & \checkmark & & \\
            AT~\cite{zagoruyko2017paying} & & &  & \\
            PKT~\cite{passalis2018learning} & & & & \\
            FT~\cite{kim2018paraphrasing} & \checkmark & \checkmark & \checkmark & \\
            DAB~\cite{heo2019knowledge} & \checkmark & \checkmark & \checkmark & \\
            RKD~\cite{park2019relational} & & & & \\
            VID~\cite{ahn2019variational} & & \checkmark & \checkmark & \\
            CCKD~\cite{peng2019correlation} & & \checkmark & \checkmark & \\
            HND~\cite{matsubara2019distilled} & & & & \\
            SPKD~\cite{tung2019similarity} & & & & \\
            CRD~\cite{tian2020contrastive} & & \checkmark & \checkmark & \checkmark \\
            Tf-KD~\cite{yuan2020revisiting} & & & & \\
            GHND~\cite{matsubara2020neural} & & & & \\
            SSKD~\cite{xu2020knowledge} & \checkmark & \checkmark & \checkmark & \checkmark \\
            $L_2$~\cite{zhang2020prime} & & & & \\
            PAD-$L_2$~\cite{zhang2020prime} & \checkmark & \checkmark & \checkmark & \\
            \hline
        \end{tabular}
    \egroup
    \end{center}
    \label{table:ref_methods}
\end{table}

\subsection{Reimplementations}
\label{subsec:reimplementations}
Given that the pretrained models in torchvision are trained on large benchmark datasets, ImageNet (ILSVRC 2012)~\cite{russakovsky2015imagenet}, and COCO 2017~\cite{lin2014microsoft}, we focus our implementations on these datasets as the pretrained models can be used as teacher models and/or baseline student models (naively trained on human-annotated datasets).
Note that some of the methods are not validated on these datasets in their original work.

Table~\ref{table:ref_methods} shows a brief summary of reference distillation methods reimplemented with \FRAMEWORK, and indicates what additional modules were implemented and added to the registry for reimplementing the methods.
We emphasize that methods without any check marks (\checkmark) in the \emph{Required additional modules} columns such as KD, AT, PKT, RKD, HND, SPKD, Tf-KD, GHND and $L_2$ can be reimplemented simply by adding the new loss modules to the registry in the framework (Section~\ref{subsec:registry}).

Different from the existing frameworks~\cite{zagoruyko2017paying,passalis2018learning,heo2019knowledge,park2019relational,tian2020contrastive,xu2020knowledge}, all the methods in Table~\ref{table:ref_methods} are reimplemented independently from models in torchvision so that users can easily switch models by specifying a model name and its parameters in a configuration file.
Taking image classification as an example, the shapes of inputs and (intermediate) outputs for the models are often fixed (\emph{e.g.}, $3 \times 224 \times 224$ and 1,000 respectively, for models trained on ImageNet dataset), that makes it easy to match the shape of student's output with that of teacher when computing loss values to be minimized.


\subsection{Reproducing ImageNet experiments}
\label{subsec:imagenet_experiments}
In this section, we attempt to reproduce some experimental results with their proposed distillation methods.
In particular, we choose the attention transfer (AT), factor transfer (FT)~\cite{kim2018paraphrasing}, contrastive representation distillation (CRD)~\cite{tian2020contrastive}, teacher-free knowledge distillation (Tf-KD)~\cite{yuan2020revisiting}, self-supervised knowledge distillation (SSKD)~\cite{xu2020knowledge}, $L_2$ and prime-aware adaptive distillation (PAD-$L_2$) methods~\cite{zhang2020prime} for the following reasons:

\begin{itemize}
    \setlength\itemsep{0em}
    \item these methods are validated with the ImageNet datasets for ResNet-34 and ResNet-18 as teacher and student models in their original work; \footnote{The teacher model for Tf-KD is the pretrained ResNet-18~\cite{yuan2020revisiting}.}
    \item the hyperparameters used in the ImageNet experiments are described in the original studies and/or their published source code; and
    \item we did not have time to tune hyperparameters for other methods that are not validated on the ImageNet dataset in their original papers.
\end{itemize}

In addition to the methods, we apply knowledge distillation (KD)~\cite{hinton14distilling} to the same teacher-student pair.
Note that except KD~\footnote{For KD, we set hyperparameters as follows: temperature $T = 1$ and relative weight $\alpha = 0.5$.}, we reuse the hyperparameters (\emph{e.g.}, number of epochs) for ImageNet given in their original work to reproduce their experimental results, and we provide the configuration and log files, and trained model weights.\textsuperscript{\ref{fn:configs}}

We also should note that Zagoruyko and Komodakis~\cite{zagoruyko2017paying} propose attention transfer (AT), and define the following total loss function for their ImageNet experiment:

\begin{equation}
    \mathcal{L}_{AT} = \mathcal{L}(\textsc{W}_{S}, x) + \frac{\beta}{2}\sum_{j \in \mathcal{I}} \norm{\frac{Q_{S}^{j}}{\norm{Q_{S}^{j}}}_{2} - \frac{Q_{T}^{j}}{\norm{Q_{T}^{j}}}_{2}}_{p},
    \label{eq:at_paper}
\end{equation}

\noindent where $\mathcal{L}(\textsc{W}_{S}, x)$ is a standard cross entropy loss, and $Q_{S}^{j}$ and $Q_{T}^{j}$ denote the vectorized forms of the $j$-th pair of student and teacher attention maps, respectively (Refer to their work~\cite{zagoruyko2017paying} for more details).
In their published framework\footnote{\url{https://github.com/szagoruyko/attention-transfer}}, they set $\beta$ and $p$ to 1,000 and 2 respectively.
However, we find a discrepancy between their defined loss function (Eq. (\ref{eq:at_paper})) and their implemented loss function (Eq. (\ref{eq:at_code})), that computes mean squared error (MSE) between the teacher and student attention maps.

\begin{equation}
    \mathcal{L}_{AT} = \mathcal{L}(\textsc{W}_{S}, x) + \frac{\beta}{2}\sum_{j \in \mathcal{I}} MSE\Bigl(\frac{Q_{S}^{j}}{\norm{Q_{S}^{j}}}_{2}, \frac{Q_{T}^{j}}{\norm{Q_{T}^{j}}}_{2}\Bigr)
    \label{eq:at_code}
\end{equation}

In our preliminary experiment with hyperparameters the authors provide, the student model did not train well with the loss module based on Eq. (\ref{eq:at_paper}).
For this reason, we used Eq. (\ref{eq:at_code}) instead for AT in our experiments.

\begin{table}[t]
    \caption{Validation accuracy of ResNet-18 (student) trained on ILSVRC 2012 dataset with ResNet-34 (teacher), using eight different distillation methods. With the hyperparameters (\emph{e.g.}, \# Epochs) either described in the original work or given by the authors, all the reimplemented methods outperform the student model trained without teacher.}
    \begin{center}
        \bgroup
        \setlength{\tabcolsep}{0.3em}
        \def\arraystretch{1.1}
        \begin{tabular}{|c|r|r|r|r|}
            \hline
            & \multicolumn{2}{c|}{\bf Accuracy[\%]} & \multirow{2}{*}{\bf \# Epochs} & \multirow{2}{*}{\bf Training time} \\ \cline{2-3}
            & \multicolumn{1}{c|}{~Top-1~} & \multicolumn{1}{c|}{Diff.} & & \\ \hline \hline
            Teacher: ResNet-34 & 73.31 & +3.56 & \multicolumn{1}{c|}{N/A} & \multicolumn{1}{c|}{N/A} \\
            Student: ResNet-18 & 69.75 & 0.00 & \multicolumn{1}{c|}{N/A} & \multicolumn{1}{c|}{N/A} \\ \hline
            KD & 71.23 & +1.48 & 100 & 60hr 04min \\
            KD~\textdagger & 71.37 & +1.62 & 100 & 23hr 07min \\
            AT & 70.90 & +1.15 & 100 & 59hr 07min \\
            AT~\textdagger & 70.55 & +0.80 & 100 & 23hr 11min \\
            FT & 71.56 & +1.81 & 91 & 55hr 06min \\
            FT~\textdagger & 71.13 & +1.38 & 91 & 22hr 15min \\
            CRD & 70.81 & +1.06 & 100 & 356hr 31min \\
            CRD~\textdaggerdbl & 70.93 & +1.18 & 100 & 179hr 12min \\
            Tf-KD & 70.52 & +0.77 & 90 & 46hr 34min \\
            Tf-KD~\textdagger & 70.21 & +0.46 & 90 & 18hr 50min \\
            SSKD~\textdaggerdbl & 70.09 & +0.34 & 130 & 113hr 12min \\
            $L_2$~\textdaggerdbl & 71.08 & +1.33 & 90 & 21hr 25min \\
            PAD-$L_2$~\textdaggerdbl & 71.71 & +1.96 & (90 +) 30 & 28hr 34min \\
            \hline
        \end{tabular}
    \egroup
    \end{center}
    \small
    \textdagger~Distributed training on 3 GPUs with linear scaling rule~\cite{goyal2017accurate}: Learning rates are modified according to the number of distributed training processes. (\emph{i.e.} multiplied by the number of GPUs). \\
    \textdaggerdbl~Distributed training on 3 GPUs with total batch size used in original work.
    \label{table:imagenet_experiments}
\end{table}

Table~\ref{table:imagenet_experiments} summarizes the results of the experiments with the training configurations (\emph{e.g.}, teacher-student pair, hyperparameters) described in each of the original studies and/or verified by the authors.
In addition to experiments with a single GPU, we perform experiments with a distributed training strategy supported by PyTorch (reported with a dagger mark \textdagger) to demonstrate that our framework supports the strategy for saving training time.
As for the $L_2$ and PAD-$L_2$ methods, the original study~\cite{zhang2020prime} uses batch size of 512 for their ImageNet experiments, which did not fit in our single GPU.
Thus, we split the batch size into 171 per GPU, and report only the results with the distributed training (marked with \textdaggerdbl).
The same strategy is applied to SSKD (total batch size of 256 and 768 for normal and augmented samples, respectively~\cite{xu2020knowledge}) as it takes at least 4 times as long at epoch-level to train a model, compared to the other methods due to their 4x augmented training data, and our batch size per GPU is 85 (for normal samples + 255 for augmented samples).
Similarly, we apply the same strategy for CRD due to the limited time.
We also note that Zhang~\emph{et al.}~\cite{zhang2020prime} applied their proposed PAD-$L_2$ to the student model trained with their proposed $L_2$ as a pretrained model, and train the student model with the PAD-$L_2$ method for 30 more epochs (\emph{i.e.}, 120 epochs).\footnote{The configuration is not described in~\cite{zhang2020prime}, but verified by the authors.}

Based on the methods we reimplemented with \FRAMEWORK, we successfully reproduce the results on the ILSVRC 2012 dataset for the teacher-student pair reported in the original papers of AT~\cite{zagoruyko2017paying}, Tf-KD~\cite{yuan2020revisiting}, $L_2$ and PAD-$L_2$~\cite{zhang2020prime} methods, and the result of PAD-$L_2$ was recently reported as the state-of-the-art performance for the teacher-student pair on the ILSVRC 2012 dataset~\cite{zhang2020prime}.
All the results outperform the baseline performance (S: ResNet-18) which is trained with human-labels only, and the pretrained model is provided by torchvision.
Note that FT was validated on ILSVRC 2015 dataset in their original work~\cite{kim2018paraphrasing}, and we confirm the FT's improvement over a baseline using ILSVRC 2012 dataset as the teacher model (ResNet-34) in torchvision is pretrained on the dataset.
The result with the reimplemented CRD is almost comparable to the accuracy reported in the original study~\cite{tian2020contrastive}.
In CRD, both positive and negative samples are leveraged for learning representations, thus turns out to be the most-time consuming method in Table~\ref{table:imagenet_experiments}.
The reimplemented SSKD outperforms the baseline model although the accuracy does not match the reported result~\cite{xu2020knowledge}.
A potential factor may be a different training configuration forced by our limited computing resource (\emph{e.g.}, different batch size per GPU whereas 8 parallel GPUs were used in their work) since we simply refactored and made the authors' published code compatible with the ILSVRC 2012 dataset.
As pointed out by Tian~\emph{et al.}~\cite{tian2020contrastive}, KD~\cite{hinton14distilling} is still a powerful method.
Our reimplmented KD outperformed their proposed state-of-the-art method, CRD (71.17\%), and achieved the comparable accuracy with their CRD+KD (71.38\%) method.

\subsection{Reproducing COCO experiments}
\label{subsec:coco_experiments}
To demonstrate that our framework can 1) be applied to different tasks, and 2) collaborate with model architectures that are not implemented in torchvision, we apply the generalized head network distillation (GHND) to bottleneck-injected R-CNN object detectors for split computing~\cite{matsubara2020neural}, using COCO 2017 dataset.
Their proposed bottleneck-injected Faster and Mask R-CNNs with ResNet-50 and FPN are designed to be partitioned into head and tail models which will be deployed on mobile device and edge server respectively, for reducing inference speed in resource-constrained edge computing systems.
Following the original work on GHND, we apply the method to a pair of the original and bottleneck-injected Faster R-CNNs as teacher and student respectively, and conduct the same experiment for Mask R-CNN as well.
As shown in Table~\ref{table:coco_experiments}, the reproduced mean average precision (mAP) match those reported in the original study~\cite{matsubara2020neural}.

\begin{table}[t]
    \caption{Validation mAP of bottleneck-injected R-CNN models for split computing (student) trained on COCO 2017 dataset by GHND with original Faster/Mask R-CNN models (teacher). Reproduced results match those reported in the original work~\cite{matsubara2020neural}.}
    \begin{center}
        \bgroup
        \setlength{\tabcolsep}{0.3em}
        \def\arraystretch{1.1}
        \begin{tabular}{|c|r|r|r|r|}
            \hline
            \multirow{2}{*}{\bf Backbone: ResNet-50 and FPN}& \multicolumn{2}{c|}{\bf mAP} & \multirow{2}{*}{\bf \# Epochs} & \multirow{2}{*}{\bf Training time} \\ \cline{2-3}
            & \multicolumn{1}{c|}{~BBox~} & \multicolumn{1}{c|}{~Mask~} & & \\ \hline \hline
            Faster R-CNN w/ Bottleneck & 0.359 & \multicolumn{1}{c|}{N/A} & 20 & 24hr 13min \\
            Mask R-CNN w/ Bottleneck & 0.369 & 0.336 & 20 & 24hr 21min \\ \hline
        \end{tabular}
    \egroup
    \end{center}
    \label{table:coco_experiments}
\end{table}



\section{Conclusions}
In this work, we presented \FRAMEWORK, an open-source framework dedicated for knowledge distillation studies, that supports efficient training and configurations systems designed to give users a summary of the experiments.
Researchers can build on the framework (\emph{e.g.}, by forking the repository) to conduct their knowledge distillation studies, and their studies can be integrated to the framework by sending a pull request.
This will help the research community ensure the reproducibility of the work, and advance the deep learning research while supporting fair method comparison on benchmarks.
Specifically, researchers can publish the log, configuration, and pretrained model weights for their champion performance, that will help them ensure the champion performance for specific datasets and teacher-student pairs.

Furthermore, the configuration files for and log files produced by \FRAMEWORK will help researchers complete the \emph{ML Code Completeness Checklist},\textsuperscript{\ref{fn:ml_code_checklist}} and we provide the full configurations (hyperparameters), log files and checkpoints including model weights for experimental results shown in Tables~\ref{table:imagenet_experiments} and \ref{table:coco_experiments} in our code repository.\textsuperscript{\ref{fn:our_repo}} 
We provide reference code and configurations for image classification and object detection tasks, and plan to extend our framework for different tasks using popular packages \emph{e.g.}, Transformers~\cite{wolf2020transformers} for NLP tasks.
Our framework will be maintained and updated along with the new releases of PyTorch and torchvision so that users can save time for coding and use it as a standard framework for reproducible knowledge distillation studies.

\section*{Acknowledgments}
We thank the anonymous reviewers for their comments and the authors of related studies for publishing their code and answering our inquiries about their experimental configurations.
We also thank Sameer Singh for feedback about naming the framework.

%
%
%
%
\bibliographystyle{splncs04}
\bibliography{references}





\appendix

\section{Hard-coded Module and Forward Hook Configurations}
\label{appendix:hardcoded}
For lowering barriers to high-quality knowledge distillation studies, it would be important to enable users to collaborate with models implemented in popular libraries such as torchvision.
However, all the models in the existing frameworks described in this study are reimplemented to extract intermediate representations in addition to the models' final outputs.
Figure~\ref{fig:model_implementations} shows an example of original and hard-coded (reimplemented) forward functions in ResNet model for knowledge distillation experiments.
As illustrated in the hard-coded example, the authors~\cite{tian2020contrastive,xu2020knowledge} unpacked an existing implementation of ResNet model and re-designed interfaces of some modules to extract additional representations (\emph{i.e.}, ``f0'', ``f1\_pre'', ``f2'', ``f2\_pre'', ``f3'', ``f3\_pre'', and ``f4'').

\begin{figure*}[t]
\centering
\noindent\begin{minipage}[t]{0.4\linewidth}
\begin{lstlisting}[language=Python, backgroundcolor=\color{MintCream}, basicstyle=\scriptsize, columns=fullflexible]
def _forward_impl(self, x):
    x = self.conv1(x)
    x = self.bn1(x)
    x = self.relu(x)

    x = self.layer1(x)
    x = self.layer2(x)
    x = self.layer3(x)

    x = self.avgpool(x)
    x = torch.flatten(x, 1)
    x = self.fc(x)

    return x

def forward(self, x):
    return self._forward_impl(x)
\end{lstlisting}
\end{minipage}\hfill
\begin{minipage}[t]{0.55\linewidth}
\begin{lstlisting}[language=Python, backgroundcolor=\color{MintCream}, basicstyle=\scriptsize, columns=fullflexible]
def forward(self, x, is_feat=False, preact=False):
    x = self.conv1(x)
    x = self.bn1(x)
    x = self.relu(x)
    f0 = x

    x, f1_pre = self.layer1(x)
    f1 = x
    x, f2_pre = self.layer2(x)
    f2 = x
    x, f3_pre = self.layer3(x)
    f3 = x

    x = self.avgpool(x)
    x = x.view(x.size(0), -1)
    f4 = x
    x = self.fc(x)

    if is_feat:
        if preact:
            return [f0, f1_pre, f2_pre, f3_pre, f4], x
        else:
            return [f0, f1, f2, f3, f4], x
    else:
        return x
\end{lstlisting}
\end{minipage}
\vspace{-1em}
\caption{Forward functions in \uline{\bf original (left, torchvision-style)} and \uline{\bf hard-coded (right, \cite{tian2020contrastive,xu2020knowledge})} implementations of ResNet. Only ``x'' from ``self.fc'' is used for vanilla training and prediction.}
\label{fig:model_implementations}
\end{figure*}

Furthermore, the modified interfaces also require those in the downstream processes to be modified accordingly, that will need extra coding cost.
We emphasize that users are required to repeat this procedure every time they introduce new models for experiments, and the same issues will be found when introducing new schemes implemented as other types of module (\emph{e.g.}, dataset and sampler) required by specific methods such as CRD~\cite{tian2020contrastive} and SSKD~\cite{xu2020knowledge}.
Using a forward hook manager in our framework, we can extract intermediate representations from the original models (\emph{e.g.}, Fig.~\ref{fig:model_implementations} (left)) without reimplementation like Fig.~\ref{fig:model_implementations} (right), and help users introduce such schemes with wrappers of the module types so that they can apply the schemes simply by specifying in a configuration file used to design an experiment.

The following example illustrates how to specify the input to or output from modules we would like to extract from ResNet model whose forward function is shown in Fig.~\ref{fig:model_implementations} (left).
``f0'', ``f1\_pre'', ``f2\_pre'', and ``f3\_pre'' in Fig.~\ref{fig:model_implementations} (right) correspond to the output from the first ReLU module ``relu'', and pre-activation representations in ``layer1'', ``layer2'', and ``layer3'' modules, which are the inputs to their last ReLU modules (\emph{i.e.}, ``layer1.1.relu'', ``layer2.1.relu'', and ``layer3.1.relu'').
``f4'' is the flatten output from average pooling module ``avgpool''.
Similarly, we can define a forward hook manager for teacher model, and reuse the module paths such as ``layer1.1.relu'' to define loss functions in the configuration file.

\begin{lstlisting}[language=yaml, backgroundcolor=\color{MintCream}, basicstyle=\scriptsize]
student:
  ...
  forward_hook:
    input: ['layer1.1.relu', 'layer2.1.relu', 'layer3.1.relu', 'fc']
    output: ['relu']
\end{lstlisting}

\section{Example PyYAML Configuration}
\label{appendix:example_config}

Figure~\ref{fig:example_config} shows an example PyYAML configuration file~\textsuperscript{\ref{fn:configs}} to instantiate abstracted modules for an experiment with knowledge distillation by Hinton \emph{et al.}~\cite{hinton14distilling}.

\begin{figure*}[!hb]
\centering
\noindent\begin{minipage}[t]{0.55\linewidth}
\begin{lstlisting}[language=yaml, backgroundcolor=\color{MintCream}, columns=fullflexible, basicstyle=\scriptsize]
datasets:
  ilsvrc2012:
    name: &dataset_name 'ilsvrc2012'
    type: 'ImageFolder'
    root: &root_dir !join ['~/dataset/', *dataset_name]
    splits:
      train:
        dataset_id: &imagenet_train !join [*dataset_name, '/train']
        params:
          root: !join [*root_dir, '/train']
          transform_params:
            - type: 'RandomResizedCrop'
              params:
                size: &input_size [224, 224]
            - type: 'RandomHorizontalFlip'
              params:
                p: 0.5
            - &totensor
              type: 'ToTensor'
              params:
            - &normalize
              type: 'Normalize'
              params:
                mean: [0.485, 0.456, 0.406]
                std: [0.229, 0.224, 0.225]
      val:
        dataset_id: &imagenet_val !join [*dataset_name, '/val']
        params:
          root: !join [*root_dir, '/val']
          transform_params:
            - type: 'Resize'
              params:
                size: 256
            - type: 'CenterCrop'
              params:
                size: *input_size
            - *totensor
            - *normalize

models:
  teacher_model:
    name: 'resnet34'
    params:
      num_classes: 1000
      pretrained: True
    ckpt: '/path/to/your_own_checkpoint_if_you_have'
  student_model:
    name: 'resnet18'
    params:
      num_classes: 1000
      pretrained: False
    ckpt: './imagenet/kd/ilsvrc2012-resnet18_from_resnet34.pt'
\end{lstlisting}
\end{minipage}\hfill
\begin{minipage}[t]{0.35\linewidth}
\begin{lstlisting}[language=yaml, backgroundcolor=\color{MintCream}, columns=fullflexible, basicstyle=\scriptsize]
train:
  log_freq: 1000
  num_epochs: 100
  train_data_loader:
    dataset_id: *imagenet_train
    random_sample: True
    batch_size: 256
    num_workers: 16
    cache_output:
  val_data_loader:
    dataset_id: *imagenet_val
    random_sample: False
    batch_size: 128
    num_workers: 16
  teacher:
    sequential: []
    wrapper: 'DistributedDataParallel'
    requires_grad: False
  student:
    sequential: []
    wrapper: 'DistributedDataParallel'
    requires_grad: True
    frozen_modules: []
  apex:
    requires: False
    opt_level: '01'
  optimizer:
    type: 'SGD'
    params:
      lr: 0.1
      momentum: 0.9
      weight_decay: 0.0001
  scheduler:
    type: 'MultiStepLR'
    params:
      milestones: [30, 60, 90]
      gamma: 0.1
  criterion:
    type: 'GeneralizedCustomLoss'
    org_term:
      criterion:
        type: 'KDLoss'
        params:
          temperature: 1.0
          alpha: 0.5
          reduction: 'batchmean'
      factor: 1.0
    sub_terms:

test:
  test_data_loader:
    dataset_id: *imagenet_val
    random_sample: False
    batch_size: 1
    num_workers: 16
\end{lstlisting}
\end{minipage}
\vspace{-1em}
\caption{First (left) and second (right) halves of an example PyYAML configuration to design a knowledge distillation experiment with hyperparameters using \FRAMEWORK.}
\label{fig:example_config}
\end{figure*}
\end{document}